%% file: main_kdd.tex
\documentclass[sigconf]{acmart}

\renewcommand\footnotetextcopyrightpermission[1]{} 
\usepackage{xspace}
\usepackage{listings}

\usepackage{caption}
\usepackage{subcaption}

\usepackage{tikz}
\usetikzlibrary{shapes,positioning}

\usepackage{pythonhighlight}

\definecolor{gooPurple}{HTML}{d9d2e9}
\definecolor{gooBlue}{HTML}{c9daf8}
\definecolor{gooGreen}{HTML}{d9ead3}
\usepackage{amsmath}
\usepackage{enumitem}
\usepackage{MnSymbol}

\usepackage{algorithm}
\usepackage{algorithmic}

\newcommand{\gnn}{TF-GNN\xspace{}} 
\newcommand{\google}{Google}
\newcommand{\rnum}[1]{\lowercase\expandafter{\romannumeral #1\relax}}
\newcommand{\coreml}{$^\mathparagraph$}
\newcommand{\research}{$^\dagger$}
\newcommand{\dm}{$^\ddag$}
\newcommand{\epfl}{$^\diamond$}
\newcommand{\topc}{$^\star$}

\AtBeginDocument{%
  \providecommand\BibTeX{{%
    \normalfont B\kern-0.5em{\scshape i\kern-0.25em b}\kern-0.8em\TeX}}}

\setcopyright{acmcopyright}
\copyrightyear{2018}
\acmYear{2018}
\acmDOI{XXXXXXX.XXXXXXX}

\acmConference[Conference acronym 'XX]{Make sure to enter the correct
  conference title from your rights confirmation emai}{June 03--05,
  2018}{Woodstock, NY}
\acmPrice{15.00}
\acmISBN{978-1-4503-XXXX-X/18/06}

\begin{document}

\title{TF-GNN: Graph Neural Networks in TensorFlow}

\author{
Oleksandr~Ferludin\topc\coreml, Arno~Eigenwillig\topc\coreml, Martin~Blais\topc\research, Dustin~Zelle\topc\research, Jan~Pfeifer\topc\coreml, Alvaro~Sanchez-Gonzalez\topc\dm, Wai Lok Sibon Li\topc\dm, Sami~Abu-El-Haija\research, Peter~Battaglia\dm, Neslihan~Bulut\research, Jonathan~Halcrow\research, Filipe~Miguel Gon{\c{c}}alves de Almeida\research, Pedro Gonnet\research,  Liangze~Jiang\epfl, Parth~Kothari\coreml, Silvio~Lattanzi\research, Andr{\'e}~Linhares\research, \mbox{Brandon} Mayer\research, Vahab~Mirrokni\research, John~Palowitch\research, Mihir~Paradkar\coreml, Jennifer~She\dm, Anton~Tsitsulin\research, Kevin~Villela\dm, Lisa~Wang\dm, David~Wong\dm, Bryan~Perozzi\topc\research
}
\email{tensorflow-gnn@googlegroups.com}
\affiliation{%
  \institution{\coreml: Google Core ML, \research: Google Research, \dm: DeepMind, \epfl: EPFL (work done at Google Research)}
  \institution{\topc: TF-GNN Top contributors}
  \country{}
}

\renewcommand{\shortauthors}{Ferludin, Eigenwillig, Blais, Zelle, Pfeifer, Sanchez-Gonzalez, et al.}


\begin{abstract}
TensorFlow-GNN (\gnn) is a scalable library for Graph Neural Networks in TensorFlow.
It is designed from the bottom up to support the kinds of rich heterogeneous graph data that occurs in today's information ecosystems.
In addition to enabling machine learning researchers and advanced developers, \gnn\ offers low-code solutions to empower the broader developer community in graph learning.
Many production models at \google\ use \gnn\, and it has been recently released as an open source project.
In this paper we describe the \gnn\ data model, its Keras message passing API, and relevant capabilities such as graph sampling and distributed training.
\end{abstract}

\received{20 February 2007}
\received[revised]{12 March 2009}
\received[accepted]{5 June 2009}

\maketitle

\section{Introduction}
Machine Learning (ML) techniques have applications across domains as varied as medicine, social networks, biochemistry, robotics, and more.
The success of many ML models is driven by their ability to incorporate different modalities of data (e.g.\ vision, text, sound, timeseries and geometric), each with its own unique structural (ir)regularities.
Traditionally, software frameworks for machine learning (e.g.\ TensorFlow \citep{abadi2016tensorflow}, PyTorch \citep{pytorch}) have focused on modeling one modality at a time, such as vision \citep{lecun98, alexnet} or natural language \citep{word2vec, transformer, bert}.
However, the development of graph representation learning \cite{chami22taxonomy,hamilton2020book} and subsequent industry interest \cite{abu2019mixhop,halcrow2020grale, kapoor2020examining, bojchevski2020scaling, al2019ddgk, rozemberczki2021pathfinder, markowitz2021graph, zhu2021robust, palowitch2022graphworld, yoon2022zeroshot} has motivated the need for better software frameworks for learning with \textit{graph}-structured data.

In this paper we introduce \gnn\footnote{\url{https://github.com/tensorflow/gnn}}, a Python framework that extends TensorFlow \citep{abadi2016tensorflow} with Graph Neural Networks (GNNs) \cite{chami22taxonomy,hamilton2020book}: models that leverage graph-structured data.
\gnn\ is motivated and informed by years of applying graph representation learning to practical problems at \google. 
In particular, \gnn\ focuses on the representation of \textit{heterogeneous} graph data and supports the explicit modeling of an arbitrary number of relationship (edge) types between an arbitrary number of entity (node) types.
These relationships can be used in combination with other TensorFlow components, e.g., a \gnn\ model might connect representations from a language model to those of a vision model and fine-tune these features for a node classification task.
Many teams at \google\ run \gnn\ models in production.
We believe this to be a direct consequence of \gnn's multi-layered API which is designed for accessibility to developers (regardless of their prior experience with machine learning).

Other software frameworks for learning from graph data have been proposed, most notably PyTorch Geometric (PyG) \citep{pyg} and Deep Graph Library (DGL) \citep{wang2019dgl}.
We differ from DGL and PyG in three main ways.
First, \gnn\ has been designed bottom-up for modeling heterogeneous graphs.  
Second, \gnn\ offers different levels of abstraction for increased modeling flexibility.
Proficient users can leverage raw TensorFlow operations for message passing, limited only by their imagination.
Intermediate users use the Keras modeling API and pre-built convolution layers, while beginner users can use the Orchestrator to quickly experiment with GNNs.
Finally, \gnn\ is programmed on top of TensorFlow.
As such, its goal is to support the many production-ready capabilities present in the TensorFlow ecosystem.

\begin{figure}[t]
    \centering
    \includegraphics{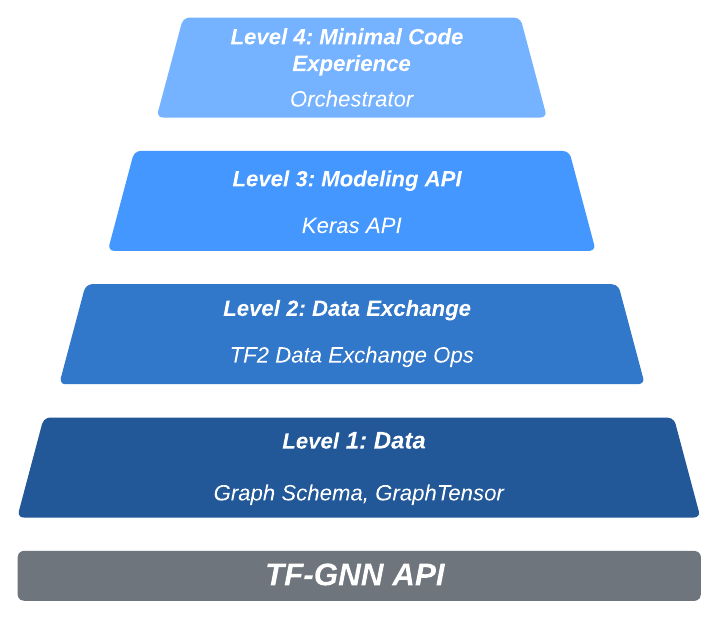}
    \caption{\gnn's layered API decomposes the tools needed to create graph models into four distinct components of increasing abstraction.}
    \label{fig:tfgnn_api}
\end{figure}

Interestingly, these differences can provide advantages for several use cases. For instance, (i)~while it is possible for PyG to model heterogeneous graph data, its syntax advocates partitioning a heterogeneous graph into a set of homogeneous graphs. 
This makes it (programmatically) challenging to create a graph layer that pools from multiple node features at once, or even create new node or feature types on the fly (i.e., through the network computation).
However, \gnn's flexibility allows for aggregating from different node or edge types at once.
Furthermore, (ii)~\gnn\ offers \textit{edge-centric}, \textit{node-centric}, and \textit{graph-centric} building blocks for GNNs. 
Existing frameworks (e.g.,\ PyG) are node-centric, making implementing some models more tedious.
On the other hand, edge-centric models, such as Graph Transformers~\cite{graphtransformer} can be natively expressed in \gnn. 
Finally, (iii)~\gnn's TensorFlow implementation inherits the benefits of the TensorFlow ecosystem, including access to model architectures for popular modalities (such as vision, text, and speech), and the ability to execute GNN models, for training and inference, on extremely fast hardware devices \citep{tpu}.

\textbf{Summary of contributions.} We present \gnn, an open-source Python library to create graph neural network models that can leverage heterogeneous relational data.
\gnn\ enables training and inference of Graph Neural Networks (GNNs) on arbitrary graph-structured data.
\gnn's four API levels allow developers of all skill levels access to powerful GNN models.
Many \gnn\ models run in production at \google.
Finally, as a native citizen of the TensorFlow ecosystem, \gnn\ shares its benefits, including pretrained models for various various modalities (e.g., a NLP model) and support for fast mathematical hardware such as Tensor Processing Units (TPUs).

\section{Overall design}
\gnn\ offers a layered API to build Graph Neural Networks that lets users trade off flexibility for abstraction. 
From least to most abstract, the layers (shown in Fig.~\ref{fig:tfgnn_api}) are:

\begin{itemize}[leftmargin=5mm]
    \item \textbf{API Level 1}: The \textit{Data Level} (\S\ref{sec:data_model}) takes care of representing heterogeneous graphs and loading them into TensorFlow, including technicalities like batching and padding.
    \item \textbf{API Level 2}: The \textit{Data Exchange Level} (\S\ref{sec:model_api_layer2}) provides operations for sending information across the graph between its \textit{nodes}, \textit{edges} and the \textit{graph context}.
    \item \textbf{API Level 3}: The \textit{Model Level} (\S\ref{sec:model_api_layer3}) facilitates writing trainable transformations of the data exchanged across the graph in order to update the state of nodes, edges and/or context.
    \item \textbf{API Level 4}: The \textit{Minimal-Code Experience Level} (\S\ref{sec:runner_api}) provides the Orchestrator,
    a toolkit for the easy composition of data input, feature processing, graph objectives, training, and validation.
\end{itemize}

This layered design is one reason for \gnn's successful adoption for graph models at \google. Users can start at a high level and only later go in deeper to tweak parts of the model. Some users may choose to only use the data level and its associated tooling (like the graph sampler, \S\ref{sec:graph_sampling}) and use their own modeling framework.
The following sections describe the API levels in greater detail. Finally, we discuss other parts of the library designed for use in production models (\S\ref{sec:gnns_at_scale}).

\section{\gnn\ Heterogeneous Data Model (API Level 1)}
\label{sec:data_model}
To train a model on heterogeneous graph data, users first need to specify its node types, edge types and their respective features. This is done with the \pyth{GraphSchema} (\S\ref{sec:graphschema}).
Based on that, the \pyth{GraphTensor} class (\S\ref{sec:graphtensor}) can represent any graphs from the dataset.

\input{images/graph_schema_example}

\subsection{Graph Schema}
\label{sec:graphschema}

A \pyth{GraphSchema} object defines:
\begin{enumerate}[leftmargin=5mm]
    \item One or more named \textbf{node sets} and their respective features.
    \item Zero or more named \textbf{edge sets} and their respective features. Each edge set has a specified
    source node set and a specified target node set. All edges in the set connect these node sets.
    \item \textbf{Context features}, which pertain to the entire input graph.\footnote{Section~\ref{sec:graphtensor} will refine the notion of context for graphs merged from a batch of inputs.}
\end{enumerate}

As the name suggests, \pyth{GraphSchema} contains only an abstract definition of how entities are related (similar to an entity relationship diagram \cite{li2009entity}) and has no actual data points. By definition, the node sets are disjoint, so they can
serve as node types; same for edges.
For each feature, the graph schema defines its name, its datatype (int, float, or string) and its shape, as in TensorFlow \citep{abadi2016tensorflow}.
That means the graph as a whole is heterogeneous, but within each node or edge set, the features are uniformly typed and shaped. Each feature can have a different shape, so the features of a node set might comprise a scalar (say, a categorical feature), a variable-length sequence (e.g., tokenized text), a fixed-length vector (such as a precomputed embedding), a rank-3 tensor with an RGB image, and so on.

\begin{figure*}
    \centering
    \includegraphics[width=\linewidth]{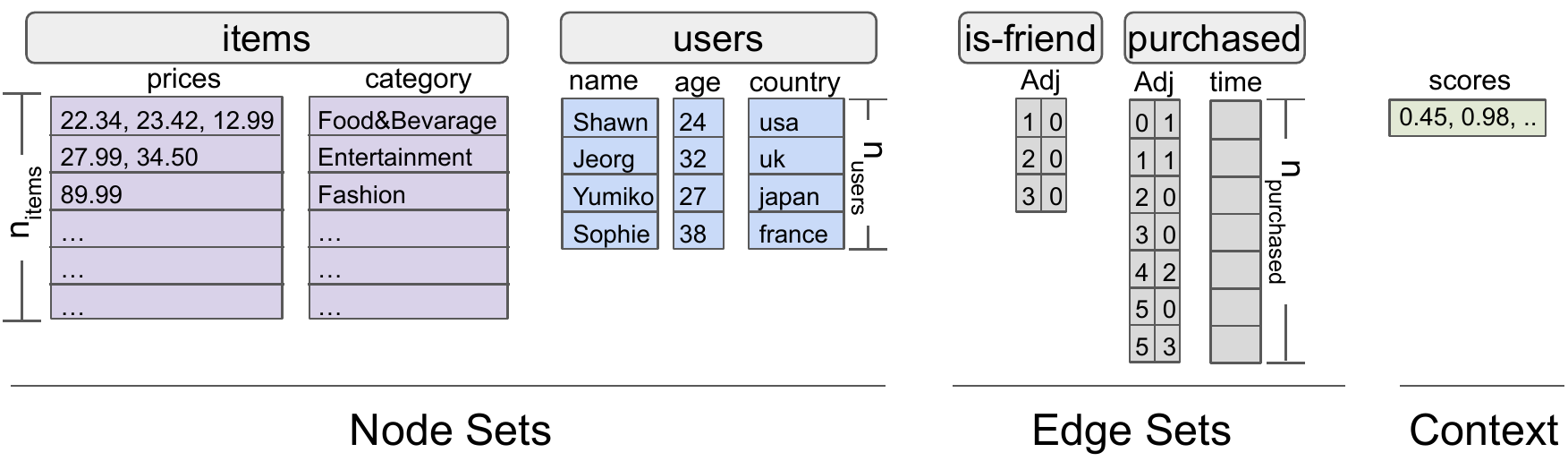}
    \caption{A \pyth{GraphTensor} to store the graph from Fig.~\ref{fig:example_graph} comprises tensors for all the features and for the adjacency data of each edge. (The size tensors are not shown here.) A Python expression for this \pyth{GraphTensor} object is shown in the Appendix.}
    \label{fig:example_graph_tensor}
\end{figure*}

An example schema for a prototypical recommendation systems problem is shown in Figure \ref{fig:example_schema_fig}.
It defines a heterogeneous graph structure with the following structure:
\begin{itemize}[leftmargin=5mm]
    \item Two node sets: ``items'' and ``users''.  \textit{Item} nodes have two features for their ``category''  (an integer  scalar, corresponding to enumeration), and their ``price'' (a floating-point vector to hold item advertised prices).
    \textit{User} nodes have three features, representing their ``name'' (\pyth{string}), ``age'' (\pyth{int}), and ``country'' (\pyth{int}).
    \item Two distinct sets of edges: ``purchased'' and ``is-friend''.  \textit{Purchased} edges connect users to items they have purchased, while \textit{is-friend} edges connect users together.
    \item One context feature ``scores'', which applies to the graph as a whole.
\end{itemize}

\subsection{GraphTensor}
\label{sec:graphtensor}

Figure~\ref{fig:example_graph} shows a graph that conforms to the example schema from above: Each circle corresponds to a node (colored by its node set), and the two types of lines correspond to the two different edge sets. Each node contains the features specified for its node set.

Figure~\ref{fig:example_graph_tensor} shows our approach to representing one such a graph in TensorFlow. We use zero-based indexing for the nodes in each node set and the edges in each edge set. Then any one feature on a node set or edge set can be represented by a tensor of shape $[n, f_1, \ldots, f_k]$, where $[f_1, \ldots, f_k]$,  $k\ge0$, is the feature shape from the GraphSchema. Moreover, for an edge set, the indices of source and target nodes are stored as two integer tensors of shape $[n]$ whose values are node indices in the node set specified by the graph schema.

The \pyth{GraphTensor} class expands on this approach to represent graphs as tensors through all stages of a TensorFlow program that builds a GNN model. Roughly, the stages are
\begin{enumerate}[leftmargin=5mm]
    \item\label{item:gt-batching} Reading, shuffling, batching and parsing GraphTensor values from \pyth{tf.Example} records on disk; possibly distributing them between replicas for data-parallel training.
    \item\label{item:gt-ftproc} Transforming one or more input features per node or edge into a fixed-size representation for deep learning.
    \item\label{item:gt-model} Running a graph neural network for several rounds to update the hidden states of nodes (and possibly edges) from neighboring parts of the graph, followed by reading out the relevant hidden states and computing the model's output.
\end{enumerate}

GraphTensor supports batching natively, and is indeed a tensor of graphs with a shape $[g_1, \ldots, g_r]$.
A scalar GraphTensor has shape $[\hspace{0.4em}]$ and holds a single graph, while a GraphTensor of shape $[g_1]$ holds a vector of $g_1$ graphs, as usual for training with minibatches. Ranks $r>1$ are rarely needed.

Each node set and edge set holds a dictionary of named features. Each feature is a tensor of shape  $[g_1, \ldots, g_r, n, f_1, \ldots, f_k]$. GraphTensor allows that a feature dimension $f_i$ may vary between the items of one node/edge set, or that the number $n$ of items varies between the multiple graphs in a GraphTensor. In both cases, the solution is to store the feature as a \pyth{tf.RaggedTensor}, not a \pyth{tf.Tensor}. Under the hood, this stores an explicit partitioning for each non-uniform, or ``ragged'', dimension. The shape reports its size as \pyth{None}.

To support feature processing, \gnn\ lets you create a new \pyth{GraphTensor} from an old one by replacing some or all of the features while keeping track of the implied schema change.

Finally, in service of training models, GraphTensor provides a method to merge a batch of inputs to a scalar GraphTensor. For each node/edge set, this method concatenates its elements across the batch of inputs and adjusts the node indices stored on edges correspondingly. The result is a GraphTensor of shape $[\hspace{0.4em}]$ with a flat index space $0, 1, \ldots, n_\mathrm{total} - 1$ for each node/edge set, across the boundaries of batched examples. Features get the shape $[n_\mathrm{total}, f_1, \ldots, f_k]$. Conveniently, such features can be represented as a \pyth{tf.Tensor} if all feature dimensions are fixed, even if a node/edge set's size $n_\mathrm{total}$ is not constant between batches. If necessary, it can be made constant as well by adding a suitably sized padding graph to each batch of input graphs and assigning it weight 0 for training the GNN.
Standard GNN operations on nodes and edges respect the boundaries between the merged input graphs, because there are no edges connecting them.
To achieve the same for context features, GraphTensor supports the notion of \textbf{components} in a graph, and stores context features indexed by component.
When a batch of graphs is merged, each input graph becomes one component of the result.

\section{Modeling with \gnn}
\label{sec:model_api}

The core of \gnn\ is specifying how a computation utilizes graph structured data.  In this section we detail the low level (TensorFlow) and high level (Keras) APIs used to construct GNNs.

\subsection{Data Exchange Ops (API Level 2)}
\label{sec:model_api_layer2}

\gnn\ sends data across the graph as follows. \textit{Broadcasting} from a node set to an edge set returns for each edge the value from the specified endpoint (say, its source node). \textit{Pooling} from an edge set to a node set returns for each node the specified aggregation (sum, mean, max, etc.) of the value on edges that have the node as the specified endpoint (say, their target node.)
The tensors involved are shaped like features of the respective node/edge set in the GraphTensor and can, but need not, be stored in it.
Similarly, graph context values can be broadcast to or pooled from the nodes or edges of each graph component in a particular a node set or edge set.

Unlike multiplication with an adjacency matrix, this approach provides a natural place to insert per-edge computations with one or more values, such as computing attention weights \citep{gat, brody2022gatv2}, integrating edge features into messages between nodes~\citep{gilmer}, or maintaining hidden states on edges~\citep{battaglia18graphnet}.

\subsection{Model Building API (API Level 3)}
\label{sec:model_api_layer3}

At API Level~3, \gnn\ follows standard TensorFlow practice and adopts Keras to express trainable transformations and their composition into models. API Levels 1 and~2 can serve other ways of modeling just as well.
The shape $[n, \ldots]$ of feature tensors allows to reuse standard neural network layers for item-wise transformations of node/edge sets, with set size $n$ in place of a batch size.

A typical GNN model (cf.~\S\ref{sec:graphtensor}) consists of (i)~feature transformations, (ii)~a base GNN, and (iii)~the final readout and prediction. \gnn\ lets you express this as a sequence of Keras layers that each take a GraphTensor input and return a GraphTensor output with transformed features -- or a Tensor for reading out the final prediction.

\subsubsection{Feature transformation layers}
\label{sec:layers_feature_xform}

The feature transformations treat each node/edge set in isolation. Depending on the available features, they can range from simple numeric transformations to running an entire deep learning model to compute an embedding of, say, image or text data. \gnn\ lets you plug in other TensorFlow models and fine-tune them jointly while training the GNN on top. In the end, the representations of multiple input features are combined to form one initial \pyth{"hidden_state"} feature. \gnn's \pyth{MapFeatures} layer makes it easy to build Keras sub-models for each node/edge set that map a features dict to a transformed features dict, and eventually the hidden state.

\subsubsection{Graph Neural Network layers}
\label{sec:layers_gnn}

The base GNN model is expressed as a sequence of \textit{graph update} layers, each of which accepts a GraphTensor with \pyth{"hidden_state"} features and returns a new GraphTensor with these features updated. Each layer object has its own trainable weights; weight sharing can be achieved by using the same layer object repeatedly.

Users can define their own graph update layers, or reuse those from the growing collection of models bundled with the \gnn\ library (e.g.,\ see the case study in Section \ref{sec:case_study}). 
A user-defined graph update can, if needed, contain free-form code with an arbitrary composition of trainable transformations and broadcast/pool operations across all parts of the graph.

More commonly, graph updates are constructed from pieces operating on individual node sets or edge sets. \gnn\ provides a generic \pyth{GraphUpdate} class for that purpose, based on the following breakdown of updating a heterogeneous graph.

Consider any node $v$ in some node set $V$ of a heterogeneous graph. Starting from the initial hidden state $\mathbf{h}_v^{(0)}$, successive \pyth{GraphUpdate}s compute the hidden state of~$v$ as
\begin{equation}
  \label{eq:next-node-state}
    \mathbf{h}_v^{(i+1)} = \textsc{NextNodeState}_V^{(i+1)}(
        \mathbf{h}_v^{(i)},
        \overline{\mathbf{m}}_{E_1,v}^{(i+1)},
        \ldots,
        \overline{\mathbf{m}}_{E_k,v}^{(i+1)})
\end{equation}
using the pooled messages $\overline{\mathbf{m}}_{E_j,v}^{(i+1)}$ received by node~$v$ in round~$i+1$ along all edge sets $E_1,\ldots,E_k$ incident to node set~$V$. Notice that their number~$k$ is a constant from the graph schema for all $v \in V$.

Let $\mathcal{N}_{E_j}(v) =\{u \mid (u,v) \in E_j\}$ denote the neighbors of $v$ along one edge set $E_j$, and notice that the size of $\mathcal{N}_{E_j}(v)$ may vary with $v \in V$. \pyth{GraphUpdate} supports two ways of computing $\overline{\mathbf{m}}_{E_j}$: in one step directly from the neighbor nodes as
\begin{equation}
  \label{eq:mbar-one-step}
  \overline{\mathbf{m}}_{E_j,v}^{(i+1)} = \textsc{Conv}_{E_j}^{(i+1)}(
      \mathbf{h}_v^{(i)},
      \{ \mathbf{h}_u^{(i)} \mid u \in \mathcal{N}_{E_j}(v)\}),
\end{equation}
or in two steps, materializing a per-edge message in the GraphTensor as
\begin{equation}
  \label{eq:mbar-two-steps}
  \begin{split}
      \mathbf{m}_{E_j,(u,v)}^{(i+1)} &= \textsc{NextEdgeState}_{E_j}^{(i+1)}(
      \mathbf{h}_u^{(i)},
      \mathbf{h}_v^{(i)},
      \mathbf{m}_{E_j,(u,v)}^{(i)}),\\
    \overline{\mathbf{m}}_{E_j,v}^{(i+1)} &= \textsc{EdgePool}_{E_j}^{(i+1)}(
      \mathbf{h}_v^{(i)},
      \{ \mathbf{m}_{E_j,(u,v)}^{(i+1)} \mid
         u \in \mathcal{N}_{E_j}(v)\}).
  \end{split}
\end{equation}
The two-step approach of Eq.~(\ref{eq:mbar-two-steps}) supports recurrence in $\mathbf{m}_{E_j,(u,v)}$, effectively turning it into a hidden state for edges. With that, and a context (or ``global'') state not shown here, this approach covers Graph Networks~\citep{battaglia18graphnet} and generalizes them to heterogeneous graphs.

Without recurrence in $\mathbf{m}_{E_j,(u,v)}$ (but possibly a constant edge feature in its place), this approach also covers Message Passing Neural Networks~\citep{gilmer}, generalized to heterogeneous graphs. The \textsc{Conv} abstraction from Eq.~(\ref{eq:mbar-one-step}) is useful to express any of a number of successful GNN architectures in a single Python class and to transfer them directly to heterogeneous graphs with an arbitrary schema. Section~\ref{sec:modelexamples} and the Appendix review some concrete cases, including Graph Convolutional Networks~\citep{gcn}, which popularized the term \emph{graph convolution} that we adopt here.

\subsection{Implementing Popular Architectures}
\label{sec:modelexamples}

\textbf{Graph Convolutional Networks} \citep{gcn}: GCNs for symmetric homogeneous graphs rely on adding loops $(v,v)$ to the single edge set $E$ to feed $\mathbf{h}_v^{(i)}$ into the computation of~$\mathbf{h}_v^{(i+1)}$ along with the neighbor nodes. \gnn\ implements them by specializing Eq.~(\ref{eq:next-node-state}) and~(\ref{eq:mbar-one-step}) to
\begin{equation}
  \label{eq:gcn}
    \mathbf{h}_v^{(i+1)} = \overline{\mathbf{m}}_{v}^{(i+1)}
  = \sigma\Big(\sum_{u \in \mathcal{N}(v) \cup \{v\}}
      \frac{1}{\sqrt{d_u d_v}} \mathbf{W}^{(i+1)} \mathbf{h}_u^{(i)}\Big),
\end{equation}
where $d_u$ is the in-degree of node $u$ including loops and $\sigma$ is an activation function, such as ReLU. Observe how \textsc{Conv} from Eq.~(\ref{eq:mbar-one-step}) is the graph convolution, and \textsc{NextNodeState} trivially passes through its result.

The relational extension, \textbf{R-GCN}, \citep{schlichtkrull2018rgcn} considers heterogeneous graphs and uses
\begin{eqnarray}
  \label{eq:rgcn-next-node-state}
  \mathbf{h}_v^{(i+1)}
  = \sigma\Big(
      \sum_{j=1}^k \overline{\mathbf{m}}_{E_j,v}^{(i+1)}
      + \mathbf{W}_V^{(i+1)} \mathbf{h}_v^{(i)} \Big),
  \\
  \overline{\mathbf{m}}_{E_j,v}^{(i+1)}
  = \frac{1}{|{\mathcal{N}_{E_j}(v)}|}
      \sum_{u \in \mathcal{N}_{E_j}(v)}
      \mathbf{W}_{E_j}^{(i+1)} \mathbf{h}_u^{(i)} \notag
\end{eqnarray}
with separate weights for each edge set and node set. This translates immediately to \textsc{Conv} and \textsc{NextNodeState} maps.

\textbf{GraphSAGE} \citep{sage} considers homogeneous sampled subgraphs and proposes several aggregator architectures, which translate directly to choices for \textsc{Conv} in Eq.~(\ref{eq:mbar-one-step}). Its \textsc{NextNodeState} function turns out to be the special case $k=1$ of Eq.~(\ref{eq:rgcn-next-node-state}) for R-GCN. In the \pyth{GraphUpdate} framework, GraphSAGE generalizes naturally to the heterogeneous case by running its \textsc{Conv}s on multiple edge sets and combining them as in Eq.~(\ref{eq:rgcn-next-node-state}) for~$k>1$.

\textbf{GAT} \citep{gat} extends GCN by replacing the weighted sum in Eq.~(\ref{eq:gcn}) by a concatenation of multiple weighted sums (attention heads), each with its own data-dependent weighting (formulas omitted for brevity). A \textsc{Conv} operation can naturally express GAT and other attention mechanisms on edges, including \textbf{GATv2}~\citep{brody2022gatv2} and \textbf{Transformer-style dot-product attention} \citep{transformer,dwivedi2021transformer,kim2022attention,brody2022gatv2}.

The \pyth{GraphUpdate} framework allows to generalize GAT/GATv2 directly to the heterogeneous case, with no extra coding, analogous to the generalization from GCN to R-GCN. Attention is distributed separately between the edges of each edge set; learning the relative importance of different edge sets (relation types) is left to their separate weight matrices~$\mathbf{W}_{E_j}^{(i+1)}$ in Eq.~(\ref{eq:rgcn-next-node-state}).

\gnn\ provides a base class for implementing \textsc{Conv} operations that allows a unified implementation of attention (i)~from a node onto its neighbors, possibly combining the neighbor node state with a feature from the connecting edge; (ii)~from a node onto its incoming edges; (iii)~from the graph context onto all nodes; (iv)~from the graph context onto all edges.
Cases~(ii--iv) provide attention for all aggregation steps of Graph Networks~\citep{battaglia18graphnet}. Appendix~\ref{appendix:gatv2} shows the unified implementation of GATv2 attention for all four cases.

\section{Orchestration (API Level 4)}
\label{sec:runner_api}

At the highest level, \gnn\ provides the \textit{Orchestrator}: a quick start toolkit with solutions for common graph learning tasks.  
It includes common graph learning objectives, distributed training capabilities, accelerator support and the handling of many TensorFlow idiosyncrasies.
The Orchestrator collects the tools necessary for (\rnum{1}) elevating the novice to a \gnn\ power user and (\rnum{2}) increasing the scope of the graph learning expert's innovation. 
The toolkit supports graph learning research by offering both a standard framework for the reproduction of results and a shared catalog of SotA and convenience graph learning techniques and objectives.
Specifically, the Orchestrator performs the:
\begin{enumerate}[leftmargin=2mm]
    \item Reading input data to extract input graph(s) as \pyth{GraphTensor} instances $\mathbf{X}$ and corresponding label(s) $\mathbf{Y}$,
    \item Processing features for a specific dataset $\mathbf{X} \rightarrow \mathbf{X}^\prime$,
    \item Adapting a model to the graph learning objective $\mathbf{M} \rightarrow \mathbf{M}^\prime$,
    \item Training the adapted model $\mathbf{M}^\prime \colon \mathbf{X}^\prime \rightarrow \mathbf{H}$ to minimize the loss between $\mathbf{H}$ and $\mathbf{Y}$,
    \item Exporting the model ($\mathbf{M}^\prime$) for inference or deployment.
\end{enumerate}

The Orchestrator provides abstractions for the composition of these five steps.
\begin{itemize}[leftmargin=2mm]
    \item \textbf{Data source} (Python class \pyth{DatasetProvider}): An arbitrary source (on disk or in memory) that produces \pyth{GraphTensor} and a corresponding schema.
    \item \textbf{Feature processing} (\pyth{GraphTensorProcessorFn}): Feature manipulations for a \pyth{GraphTensor}--typically associated with a specific dataset.
    \item \textbf{Task} (Python class \pyth{Task}): Defines the graph learning objective on top of the base GNN model (e.g., node classification, or regression, or a self-supervised task like DeepGraphInfomax \cite{dgi}) and its ancillary pieces like source data transformations.    \item \textbf{Model}: An arbitrary model that operates on \pyth{GraphTensor}. The model is adapted to a graph learning objective by the above \textbf{Task}.
    \item \textbf{Training Hyperparameters}: Including the choice of optimization algorithm (e.g., Adam \citep{adam}), learning rate, as well as model hyperparameters (e.g., number of layers and their widths). Integrated with an automated hyperparameter tuning service~\citep{golovin2017vizier,song2022ossvizier}.
\end{itemize}

\section{Sampling and Scaling}
\label{sec:gnns_at_scale}
At \google, we need to build neural network models for graphs of incredible size (trillions of edges). 
Heterogeneous graphs of this scale that have rich node and edge features cannot fit in the memory of single machines. 
\gnn\ offers a variety of approaches for transforming graph data.  In our experience, the right approach typically depends on the graph's size.
Here we discuss some details regarding processing  datasets.

\begin{figure}[!t]
    \centering
    \includegraphics[width=\linewidth]{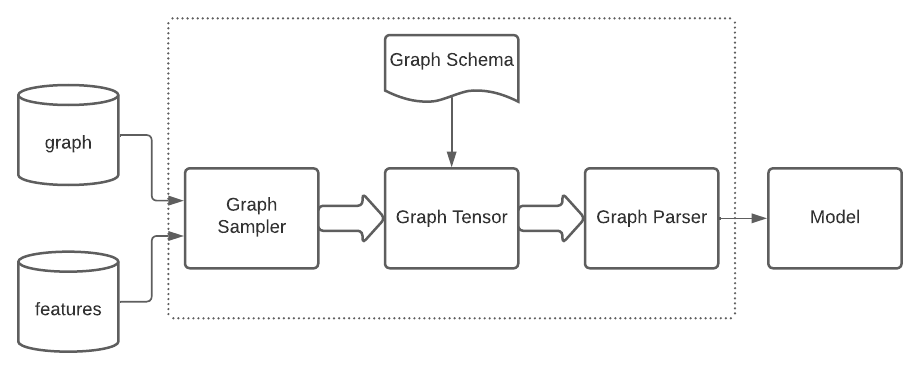}
    \vspace{-10pt}    
    \caption{Diagram of massive-graph sampling and training pipeline with \gnn.}
    \label{fig:sampling}
\end{figure}

\subsection{Sampling}
On massive, well-connected graphs, even deep GNNs rarely aggregate features from distant nodes. As a result, when making a prediction on a root node (or a root activation node for edge/neighborhood prediction), it is frequently equivalent for the GNN to operate on a sufficiently-wide \emph{subgraph} around the root~\cite{sage}. For example, training and inference with a 2-layer GCN \cite{gcn} only requires the 2-hop subgraph around the root.

\gnn\ formalizes this idea of rooted subgraph sampling with a \emph{sampling plan}.  The sampling plan describes the set of paths to sample from a graph schema in order to created rooted subgraphs.
The key concepts of a sampling plan are: (1) the \texttt{seeds} -- the nodes to start the expansion from, (2) the edge sets to expand out through, and (3) the kind of sampling to use in the expansion.
This is discussed in detail during our case study in Section \ref{sec:case_study}.

\subsubsection{Large Scale: Distributed Sampling}
\label{sec:graph_sampling}
One practical setting is the case of training GNNs on large scale data (e.g.\ graphs with \emph{billions} of nodes).
To train GNNs on massive graphs, we first run a distributed \emph{subgraph sampling} operation which queries each root node for a subgraph, and writes each subgraph to disk as an individual GraphTensor. 

The distributed sampling algorithm is presented in Algorithm~\ref{alg:sampling}.
At a high level, it operates by repeatedly applying a distributed \texttt{Sample} function to grow the effective frontier of all node's samples at once.
Each invocation of this function joins together a set of nodes to sample, and an edge set to sample over.
After performing repeated sampling operations, the samples are joined by their sample id to form connected subgraphs.
These subgraphs are then joined with the appropriate node features and converted to GraphTensors.
Figure \ref{fig:sampling}  shows an illustration of how sampling fits into the training pipeline.

The set of sampled GraphTensors is typically stored on distributed cloud storage.
For the purpose of easing downstream processing, the  GraphTensors are randomly grouped into file shards,
which are then used as input for model training or inference.

\begin{algorithm}[t]
\caption{Distributed Sampling}
\label{alg:sampling}
\begin{algorithmic}
\REQUIRE $S_\emptyset$: the initial set of nodes to sample\\ 
$E$: the edge sets to sample from\\ 
$p$: a sampling plan.
\ENSURE $\mathcal{G}$, a set of Graph Tensors\\
\hrulefill
\STATE frontier$_0$ = \texttt{Sample}($S_\emptyset$, $E_{p_0}$)
\FOR{$i = 1; i \leq p\text{.steps}$}
\STATE frontier$_i$ = \texttt{Sample}(frontier, $E_{p_i}$)
\ENDFOR

\STATE edge\_groups = frontier.GroupBy(frontier.sample\_id)

\STATE edge\_groups = DeduplicateNodes(edge\_groups)

\STATE edges\_with\_features = lookup\_features(edge\_groups)

\STATE $\mathcal{G}$ = create\_graph\_tensors(edges\_with\_features)
\end{algorithmic}
\end{algorithm}

\subsubsection{Medium Scale: In-memory Sampling}
Not all datasets are large enough to require distributed computation.
In the case of smaller datasets (perhaps those with less than $100$ million nodes), the input graphs are small enough to fit into memory on one machine.
However it can still be beneficial to perform sampling, in order to speed up training.  
In this case  all graph nodes, edges, and features, are loaded for processing at once, and an in-memory sampling operation follows the sampling plan to generate GraphTensors on-the-fly.
Here, one can express the objective function on the entire graph, e.g., as a node classification cross-entropy loss on all labeled nodes instead of piece-wise on rooted subgraphs.
\gnn\ offers separate functionality and tutorials for these settings.

We note that unlike with distributed sampling, samples generated by the in-memory procedure are typically not persisted, and instead are used on-demand during the training or inference process.

\subsubsection{Small Scale: No sampling}

There are also many applications of \gnn\ where the graph(s) under consideration are quite small, and so each graph can just be completely contained inside a single GraphTensor.
This setting is similar to the standard in-memory application of GNNs, and therefore we don't devote much space to it here.

\subsection{Training}
After creating GraphTensors, the next step is deciding how to train the model.
\gnn's orchestration layer aims to make GNN training no different than normal model training.
Specifically, it allows developers to provide a TensorFlow \texttt{tf.distribute.Strategy}\footnote{\url{https://www.tensorflow.org/guide/distributed_training}} which allows both distributed training and accelerator (GPU, TPU) use.

\subsubsection{Distributed Input Processing}

\gnn's orchestration layer also integrates with TensorFlow's \texttt{tf.data} service\footnote{\url{https://www.tensorflow.org/api_docs/python/tf/data/experimental/service}}, which allows utilizing a separate distributed compute cluster just for creating input for training, and performing CPU-based preprocessing.
This can greatly accelerate \gnn\ model training by reducing I/O bottlenecks.

\subsubsection{Model Serialization}

After the training process is complete, a standard TensorFlow \texttt{SavedModel}\footnote{\url{https://www.tensorflow.org/api_docs/python/tf/saved_model}} is created for the trained model.

\subsection{Inference}
Once we have a trained model, there are a variety of different options for performing inference.
In the simplest case, we can do offline inference with batch processing using the SavedModel and a set of serialized GraphTensors (perhaps made through the same sampling process).
More complex productionization settings can host the SavedModel on an inference service using TensorFlow Serving\footnote{\url{https://www.tensorflow.org/tfx/tutorials/serving/rest_simple}}, or convert it to TFLite for inference on a mobile or edge device.\footnote{\url{https://www.tensorflow.org/lite}}
In this case, the GraphTensors used for input will have to be generated by the calling application (perhaps via the in-memory sampler).

\section{Related Work}

Here we provide an overview of related work, which we
separate into two broad categories: single-machine and distributed software frameworks.
While it is more common for single-machine frameworks to be on the limits of research and innovation, industrial applications require distributed support and pose challenges rarely addressed by pure research applications.

\paragraph{Single-machine libraries.}

PyG (PyTorch-Geometric)~\cite{pyg} is a de-facto standard framework for GNNs in the PyTorch ecosystem.
PyG provides automatic batching support, GPU acceleration, and an interface to common graph learning datasets.
Its performance is further enhanced by a set of optimized sparse GPU kernels tailored towards graph learning workloads.
Spektral~\cite{grattarola2021spektral} is a prominent TensorFlow framework that follows Keras model building principles.
It offers a similar experience to PyG in the TensorFlow ecosystem but without any batching support.

\gnn\ differs from these in many ways.
For example, Spektral computes edge-centric functions (such as GAT) on minibatches by converting sparse adjacency matrices to dense ones.
In general, these frameworks are not designed to scale to large graphs, but allow for easy experimentation by researchers on small graphs. On the other hand, \gnn's primary purpose is to scale to large graphs.

\paragraph{Distributed libraries.}

Deep Graph Library (DGL)~\cite{wang2019dgl}
allows switching the backend platform between PyTorch, TensorFlow, and Apache MXnet with minimal code modifications. 
The DistDGL extension~\cite{zheng2020distdgl} enables efficient multi-machine training with DGL. Training is supposed to be done on a fleet of CPU-heavy instances connected in a cluster with a wide communication channel. DistDGL is mainly focused on reducing communication between the workers, each of which is performing sampling and training simultaneously. It partitions the input graph with METIS~\cite{karypis1998fast} and uses each partition as an example. 
Unlike DistDGL, \gnn's distributed training is more general.  
\gnn\ does not assume that the the data contains clusters, or that the graph structure fit into memory for partitioning via METIS.

Graph-Learn (formerly AliGraph)~\cite{zhu2019aligraph} is an open-source industrial graph learning framework built on top of TensorFlow.
It is designed to natively handle large heterogeneous graphs, and it employs several techniques to facilitate large-scale training.
Their distribution strategy relies on distributing the graph among worker machines, with a requirement that all worker machines must be alive at the same time: their training would stop if any worker machine fails. This differs from the distribution strategy of TF-GNN. In particular, TF-GNN samples a large graph into subgraphs using a resilient distributed system \citep{chambers2010flumejava}.
Similarly, \gnn\ can be used with the asynchronous distributed model training in TensorFlow, which is robust to machine failures.

Paddle Graph Learning (PGL)~\cite{ma2019paddlepaddle} is probably the most similar to \gnn. It is founded upon message passing over heterogeneous graphs.
There are two notable differences between PGL and TF-GNN. First, PGL is more restrictive: each node must have a single feature (it is non-trivial to combine visual feature and textual feature, per node, per se) and dictates that all nodes must have the same feature dimensions. In contrast, TF-GNN support multiple features per node type (including ragged feature dimensions) and two node types can hold different features. Second, PGL uses Paddle as the computation backend whereas TF-GNN uses TensorFlow. 
From a practical sense, it is easier to find state-of-the-art (SotA) network architectures and pre-trained models in TensorFlow, which would make it easier to combine per-modality SotA models within a GNN.

\section{Example Use Case: OGBN-MAG}
\label{sec:case_study}
Now we turn our attention to an end-to-end example illustrating how the components of TF-GNN interoperate within a `real world’ setting.
As most applications of TF-GNN are proprietary, in this section we consider how a hypothetical developer at Google would use TF-GNN to solve a node classification problem using an open source dataset instead (OGBN-MAG).
While the application here is admittedly contrived, the process is very illustrative of how TF-GNN is used internally.

\subsection{Problem Identification}

Our Googler begins their TF-GNN journey by identifying a product need.  
After consultation with their manager, team, and project stakeholders they determine that they desire to perform a node classification prediction for academic papers to determine the venue (journal or conference) in which each of the papers has been published.
Having identified the machine learning task at hand, they collect labels for the training dataset and  split it into three segments based on the time the papers were written ("train" (year<=2017), "validation" (year==2018), and "test"(year==2019)).

The next step in their TF-GNN journey is deciding on the appropriate graph structure to use for this classification problem.
For this step, our Googler talks to data owners and determines what kinds of entities and relations are available for use.
They refine this down to four kinds of nodes (papers, authors, institutions, and fields of study) and five kinds of edges to include in their model.
\footnote{We note that there are not too many node features included in this model, but want to make clear that in most real world settings there are an abundance of node (or edge) features possible.}
To close out this step, they encode these relationships in a Graph Schema (shown in Figure \ref{fig:case_study_schema}.
The full protocol buffer definition of the schema is available in the Appendix A7.1.

\begin{figure}
    \centering
    \vspace{-5pt}
    \includegraphics[width=\linewidth]{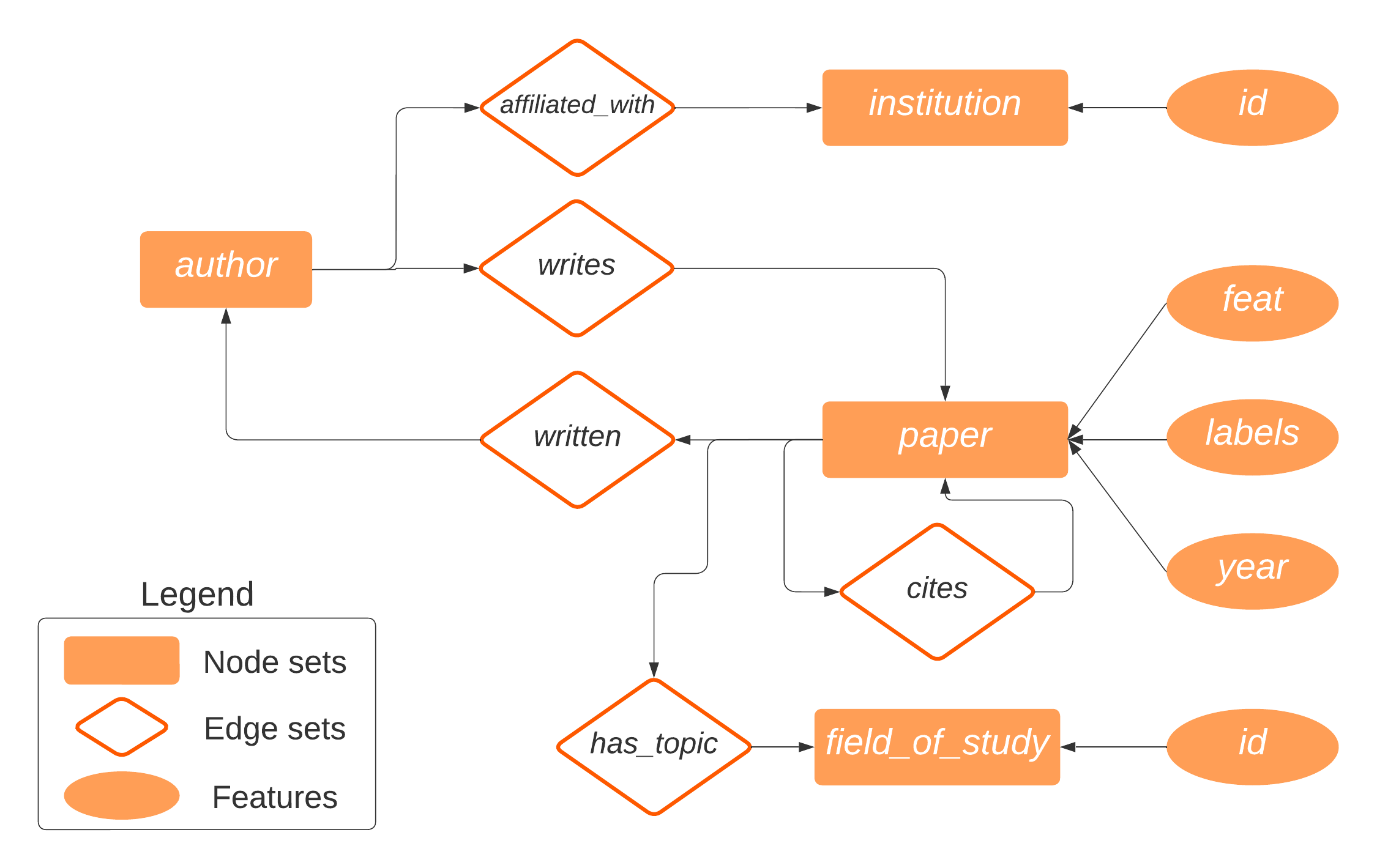}
    \caption{A Graph Schema chosen for papers classification on OGBN-MAG.  It contains 4 node types and 5 edge types.}
    \label{fig:case_study_schema}
\end{figure}

This graph schema provides rich neighborhoods to author and paper, the two most numerous kinds of nodes, from which representations can be computed. Institutions and fields of study on the other hand have an unique `id’ feature that gets embedded and forms their node representation. This will allow the GNN model to train embedding tables for their representations over time, instead of computing a representation for them on the fly from their neighbors.

\subsection{Input Generation}

Next our Googler creates a sampling specification to encode how they would like to convert the graph for their model.

Since this is a node classification problem for papers, the user chooses to extract subgraphs, each subgraph centered around a paper node. For each seed paper node, the user wishes to sample other papers the seed paper cites, authors of cited papers, their affiliations, and other papers written by them. For all sampled papers (seed, cited, authored), user wants to also sample some fields of study. This sampling logic can be concisely written using TF-GNN’s SamplingSpecBuilder – see Algorithm~\ref{alg:samplingspecbuilder}.  
The output of this sampling specification is encoded in a protocol buffer (Appendix A.7.2).
\begin{figure}
    \centering
    \begin{lstlisting}[frame=single]
# Each paper node is a seed for graph sampling.
seed_paper = SamplingSpecBuilder(schema,
                                 RANDOM_UNIFORM).seed(
                                 'paper')
# From each seed paper, sample cited papers.
cited_papers = seed_paper.sample(32, 'cites')
# From each paper (seed/cited), sample up to 8 authors.
authors = cited_papers.join([seed_paper]).sample(8, 
                                            'written')
# From these authors, sample up to 16 extra papers written by each.
author_papers = authors.sample(16, 'writes')
# From these authors, sample their affiliations.
_ = authors.sample(16, 'affiliated_with')
# From each paper (seed/cited/written), sample topics.
_ = author_papers.join([seed_paper,
                        cited_papers]).sample(16, 
                        'has_topic')
# Create the SamplingSpec protocol buffer.
sampling_spec = seed_paper.build()
    \end{lstlisting}
    \vspace{-10pt}        
    \caption{Sampling Specification for OGBN-MAG}
    \label{alg:samplingspecbuilder}
\end{figure}

From the \pyth{SamplingSpec} that gets generated by Algorithm \ref{alg:samplingspecbuilder}, our Googler user can choose which sampling pipeline to utilize for training a GNN.
In this case, let us suspend disbelief momentarily and say the user decides that their graph data is too large for a single machine -- as this is commonly the case for real datasets at Google.
Therefore, they run the distributed sampler (described in Section \ref{sec:gnns_at_scale}) and generate a set of samples from the nodes in their graph.

\subsection{Modeling}
The next step in our Googler’s journey is perhaps the most exciting part -- the modeling phase!  Here, the user must choose graph convolutions to transform data over their sampled input graphs.

After making some assumptions about the kind of information transfer necessary for the problem (and perhaps after performing a consultation with the TF-GNN team), our user decides that a simple message passing neural network (MPNN) \cite{gilmer} is a promising first approach to their problem.

For exposition, let us start by writing an MPNN from scratch, before we look at the ready-to-use model pieces shipped with TF-GNN. The two key pieces are (i) a Conv layer, to define how messages are computed on one edge set and aggregated at its receiver nodes, and (ii) a NextState layer, to define how node states are updated based on the aggregated messages from the various incoming edge sets. Figure \ref{alg:case_study_mympnn} shows the code to define a message passing neural network, with some boilerplate omitted.  
As you can see, there are 4 rounds of message passing, each implemented by a GraphUpdate layer that breaks it down into convolutions on the various edge sets and node state updates for those node sets that are not backed by embedding tables. This compact yet explicit representation of all trainable model pieces allows our user a great deal of control, separately for each edge set and node set.

\begin{figure}
    \centering
    \begin{lstlisting}[frame=single]
class MyConv(tf.keras.layers.Layer):
 """Sends messages on an edge set and pools them for each receiver node."""
 def __init__(self, units, **kwargs):
   super().__init__(**kwargs)
   self._message_fn = tf.keras.layers.Dense(units, activation="relu")

 def call(self, graph, *, edge_set_name):
   # NOTE: The receiver is the source node from which the edge was sampled.
   receiver_states = tfgnn.broadcast_node_to_edges(
       graph, edge_set_name, tfgnn.SOURCE,
       feature_name=tfgnn.HIDDEN_STATE)
   sender_states = tfgnn.broadcast_node_to_edges(
       graph, edge_set_name, tfgnn.TARGET,
       feature_name=tfgnn.HIDDEN_STATE)
   message_inputs = tf.concat([sender_states, receiver_states], axis=-1)
   messages = self._message_fn(message_inputs)
   pooled_messages = tfgnn.pool_edges_to_node(
       graph, edge_set_name, tfgnn.SOURCE, "sum", feature_value=messages)
   return pooled_messages

class MyNextNodeState(tf.keras.layers.Layer):
 """Computes next node state from pooled messages and previous node state."""
 def __init__(self, units, **kwargs):
   super().__init__(**kwargs)
   self._update_fn = tf.keras.layers.Dense(units, "relu")

 def call(self, inputs):
   prev_node_state, edge_set_inputs, context_input = inputs
   assert not context_input, "Unused in this example."
   combined_inputs = tf.concat([
       prev_node_state
   ] + [
       pooled_messages
       for edge_set_name, pooled_messages in edge_set_inputs.items()
   ], axis=-1)
   next_node_state = self._update_fn(combined_inputs)
   return next_node_state

def model_fn(graph_tensor_spec):
 """Creates the GNN model from an input spec."""
 hidden_state_dim = 128
 message_dim = 128
 input_graph_tensor = tf.keras.layers.Input(type_spec=graph_tensor_spec)
 graph_tensor = tfgnn.keras.layers.MapFeatures(
     node_sets_fn=set_initial_node_states)(input_graph_tensor)
 for _ in range(4):
   graph_tensor = tfgnn.keras.layers.GraphUpdate(node_sets={
       "paper": tfgnn.keras.layers.NodeSetUpdate(
           {"cites": MyConv(message_dim),
            "written": MyConv(message_dim),
            "has_topic": MyConv(message_dim)},
           MyNextNodeState(hidden_state_dim)),
        "author": tfgnn.keras.layers.NodeSetUpdate(
           {"writes": MyConv(message_dim),
            "affiliated_with": MyConv(message_dim)},
           MyNextNodeState(hidden_state_dim)),
        })(graph_tensor)
 return tf.keras.Model(input_graph_tensor, graph_tensor)

    \end{lstlisting}
    \vspace{-10pt}        
    \caption{MPNN implementation via GraphUpdate}
    \label{alg:case_study_mympnn}
\end{figure}

The likely next step for our user is to add dropout, weight regularization and layer normalization to the MPNN shown above. This is straightforward to add to the Keras classes shown above, albeit verbose. This is where TF-GNN’s bundled model collection comes in handy: The code in Figure \ref{alg:case_study_mympnn}  can be replaced by the much shorter Figure \ref{alg:case_study_vanilla_mpnn}.

\begin{figure}
    \begin{lstlisting}[frame=single]
from tensorflow_gnn.models import vanilla_mpnn

def model_fn(graph_tensor_spec):
  ...  # As above.
 for _ in range(4):
    graph_tensor = vanilla_mpnn.VanillaMPNNGraphUpdate(
        units=..., message_dim=..., receiver_tag=tfgnn.SOURCE,
        l2_regularization=..., dropout_rate=...,
        use_layer_normalization=True,
    )(graph_tensor)
 return tf.keras.Model(input_graph_tensor, graph_tensor)    
    \end{lstlisting}
    \vspace{-10pt}    
    \caption{MPNN reuse from model library}
    \label{alg:case_study_vanilla_mpnn}
\end{figure}

Beyond VanillaMPNN, TF-GNN offers models such as GraphSAGE, GATv2 and (Transformer-style) MultiHeadAttention. These can be used wholesale, analogous to VanillaMPNN above, or mixed and matched per edge set / node set by combining them in one GraphUpdate.

After finishing a modeling implementation (either by writing one, or using a prebuilt one), our Googler is ready to set it up for training.

\subsection{Orchestration}

In order to add a loss to the model and set up training, our Googler then turns to writing a small configuration using TF-GNN’s orchestration layer.  This involves loading the data, choosing a Task for the model (RootNodeMulticlassClassification), and running the Keras trainer. Appendix~\ref{appendix:minimal-runner-ogbnmag} shows a minimal but working TensorFlow training script that plugs the aforementioned code pieces into the \pyth{runner.run()} entry point.

Not shown here is the Runner’s support for validation, checkpointing, padding inputs to fixed sizes (as required for Cloud TPUs), customizable export to a SavedModel for inference, model attribution utilities, unsupervised graph objectives and more. Please refer to the more comprehensive \label{sec:runner_api} for detailed information.

\subsection{Model Tuning}

Finally, our Googler is ready to tune hyper-parameters to find the best model for their task.  
TF-GNN’s support for Vizier \cite{golovin2017vizier, song2022ossvizier} hyper-parameter optimization allows them to explore different parameters including model parameters such as the dimensionality of the messages used and regularization parameters like dropout.


The Googler then conducts an extensive vizier study (here 100 trials where each trial is run for 20 epochs) with the objective is to maximize the accuracy on OGBN-MAG’s validation set.  Full details of the hyper-parameter search settings, along with the results, are found in Appendix~\ref{appendix:hyper-params-optimization}.
Happily, they find a well-performing model with performance characteristics that meet their product requirements.  

Table \ref{tab:case_study_results} shows how the the results from the Googler's Vizier study compare to a model with much more capacity (the Heterogeneous Graph Transformer \cite{hgt}) from the OGBN-MAG Leaderboard.
Interestingly, we see that a simple model (with no customization) and some hyper-parameter optimization can out compete a much more complicated model utilizing transformer-style attention.
The TF-GNN platform aims to minimize the hurdles to placing these (and much more complicated) graph models into production settings.

\begin{table}
    \centering
    \begin{tabular}{c|c|c|c}
          & \# params & validation & test \\
         HGT (leaderboard) & 26.8M &  0.5124 & 0.4982\\
          & &  $\pm$ 0.0046 & $\pm$ 0.0013\\
          \hline
         MPNN (tf-gnn) & \textbf{5.89M} & \textbf{0.5149} & \textbf{0.5027}\\
         & & $\pm$ \textbf{0.0019} & $\pm$ \textbf{0.0022} \\
    \end{tabular}
    \caption{The `simple' model found here through hyper-parameter optimization outperforms a much higher capacity model from the OGB leaderboards.}
    \label{tab:case_study_results}
\end{table}

\section{Conclusions}
In this work we have presented \gnn, our open source framework used for Graph Neural Networks at \google.
\gnn\ is a software framework which reduces the technical burden for GNN productization and facilitates experimentation with GNNs.
\gnn's expressive modeling capability allows complex relationships between nodes, edges, and graph-level elements in a model.
This enables the straightforward implementation of intricate models.
\gnn\ offers four levels of increasingly abstract APIs, serving a range of use-cases from beginners with a graph problem but little ML experience to ML researchers who desire complete control of their graph learning system.
Many models at \google\ already use \gnn, and we believe that this project will accelerate the industrial adaptation of these promising models at more organizations.

\bibliographystyle{ACM-Reference-Format}
\bibliography{references}


\clearpage

\onecolumn
\appendix

\section{Appendix}

\subsection{Graph Tensor Example}
\label{appendix:graph_tensor_example}
Consider again the example schema and graph provided in Figures~\ref{fig:graph_schema_example_graph} and~\ref{fig:example_graph_tensor}.
This example graph would be represented with the following tensors:

\begin{itemize}[leftmargin=4mm]
    \item Node feature tensors:
    \begin{itemize}[leftmargin=4mm]
        \item ``items'' category, string vector tensor: \pyth{["food", "show ticket", "shoes", "book", "flight", "groceries"]}, with shape \pyth{[6]}
        \item ``item'' price, float tensor: \pyth{[[22.34, 23.42, 12.99], [27.99, 34.50], [89.99], [24.99, 45.00], [350.00], [45.13, 79.80, 12.35]]}, with shape \pyth{[6, None]}; an instance of \pyth{tf.RaggedTensor}
        \item ``users'' name, string vector tensor: \pyth{["Shawn", "Jeorg", "Yumiko", "Sophie"]}, with shape \pyth{[4]}
        \item ``users'' age, int vector tensor: \pyth{[24, 32, 27, 38]}, with shape \pyth{[4]}.
        \item ``users'' country: int vector tensor: \pyth{[3, 2, 1, 0]}, with shape \pyth{[4]}, assuming that the country vocabulary enumeration is \pyth{dict(france=0, japan=1, uk=2, usa=3)}.
    \end{itemize}
    \item Edge feature tensors:
    \begin{itemize}[leftmargin=4mm]
        \item ``purchased'' with
        \begin{itemize}[leftmargin=4mm]
            \item source indices, int vector tensor: \pyth{[0, 1, 2, 3, 4, 5, 5]}.
            \item target indices, int vector tensor: \pyth{[1, 1, 0, 0, 2, 3, 0]}, both with shape \pyth{[7]}.
        \end{itemize}
        \item ``is-friend'' with
        \begin{itemize}[leftmargin=4mm]
            \item source indices, int vector tensor: \pyth{[1, 2, 3]}.
            \item target indices, int vector tensor: \pyth{[0, 0, 0]}, both with shape \pyth{[3]}. 
        \end{itemize}
    \end{itemize}
    \item Graph-level feature tensors:
    \begin{itemize}[leftmargin=4mm]
        \item ``scores'', float matrix tensor: \pyth{[[0.45, 0.98, 0.10, 0.25]]} with shape \pyth{[1, 4]}.
    \end{itemize}
\end{itemize}

Note how the edges connectivity is encoded as source and target indices in the arrays of node features. The indices are indicating, for each edge, which position in the node feature array they are referring to. For example, the fifth values of the `purchased/\#source` and `purchased/\#target` is `[4, 2]`, which link together nodes `"flight"` and `"Yumiko"`.

\subsubsection{GraphTensor Features \& Shapes}
\label{sec:graphfeatsshapes}
Shapes of feature tensors, stored in \pyth{GraphTensor}, are crucial for building models in \gnn. 
Each tensor has a particular shape constraint based on its containing node set, edge set, or context dictionary. The shapes are as follows:
\begin{itemize}
    \item Node features. All the features associated with a node set share the initial dimension, which is the total number of nodes in the node set.
    In the example above, features for node ``items'' have 6 nodes, and so all their tensor shapes are of the form \pyth{[6, ...]}. In general, node features will have the \pyth{[num_nodes, feature...]} shape.  For example, a simple 64-dimensional embedding of each node results in shape \pyth{[6, 64]}, while a 224x224 image with 3 color channels stored at each node could result in \pyth{[6, 224, 224, 3]}.
    \item Edge features and indices. Similarly, all the features associated with an edge set shares the leading dimension, which is the total number of edges in the edge set.
    This includes the edge indices (rendered above as special features \pyth{source} and \pyth{target}). In the example, features for edges ``purchased'', both of these have shape \pyth{[7]}. If edges have information encoded as features, \textit{e.g.}, an embedding of shape \pyth{[32]}, then the edge feature tensor would be of shape \pyth{[7, 32]}.  In general, edge features will have shape \pyth{[num_edges, feature...]}.
    \item Context features. Context features apply to one component of the graph. An input graph parsed with a GraphSchema has a single component, so its context features have shapes \pyth{[1, ...]}. After batching inputs and merging batches of inputs into components of a single graph (see \S\ref{sec:graphtensor}), there are multiple components, and their number appears as the outermost dimension in the shape of context features.
\end{itemize}

\subsection{Graph Tensor API}

\subsubsection{GraphTensor Interface}

The \pyth{GraphTensor} objects you obtain from the parser are lightweight containers for all the dense and ragged features that are part of an example graph, as well as the adjacency information. These can contain a single example graph or a batch of multiple graphs.
\noindent You can access the tensors with an interface similar to that of Python dicts. For example, to access the age feature in the example, you would do this:

\begin{python}
graph.node_sets["users"]["age"]

<tf.Tensor: shape=(4,), dtype=int32, numpy=array([24, 32, 27, 38], dtype=int32)>
\end{python}

\noindent Edge indices are accessed with their "adjacency" property:

\begin{python}
graph.edge_sets["purchased"].adjacency.source

<tf.Tensor: shape=(7,), dtype=int32, numpy=array([0, 1, 2, 3, 4, 5, 5], dtype=int32)>
\end{python}

\noindent And similarly for context features:

\begin{python}
graph.context["scores"]

<tf.Tensor: shape=(1, 4), dtype=float32, numpy=array([[0.45, 0.98, 0.1 , 0.25]], dtype=float32)>
\end{python}

\subsubsection{Creating GraphTensors}


Instances of \pyth{GraphTensor} can also be created as constants or from existing tensors. This is useful for writing unit tests and working in a Colab. To create a \pyth{GraphTensor}, you have to provide the various pieces and features forming the \pyth{GraphTensor}. This is best described by an example:

\begin{python}
graph = tfgnn.GraphTensor.from_pieces(
    context=tfgnn.Context.from_fields(
        features={
            "scores": [[0.45, 0.98, 0.10, 0.25]],
        }),
    node_sets={
        "items": tfgnn.NodeSet.from_fields(
            sizes=[6],
            features={
                "category": ["food", "show ticket", "shoes", 
                             "book", "flight", "groceries"],
                "price": tf.ragged.constant([[22.34, 23.42, 12.99], 
                                             [27.99, 34.50], 
                                             [89.99], 
                                             [24.99, 45.00], 
                                             [350.00], 
                                             [45.13, 79.80, 12.35]]),
            }),
        "users": tfgnn.NodeSet.from_fields(
            sizes=[4],
            features={
                "name": ["Shawn", "Jeorg", "Yumiko", "Sophie"],
                "age": [24, 32, 27, 38],
                "country": ["usa", "uk", "japan", "france"],
            }),
    },
    edge_sets={
        "purchased": tfgnn.EdgeSet.from_fields(
            sizes=[7],
            features={},
            adjacency=tfgnn.Adjacency.from_indices(
                source=("items", [0, 1, 2, 3, 4, 5, 5]),
                target=("users", [1, 1, 0, 0, 2, 3, 0]),
            )),
        "is-friend": tfgnn.EdgeSet.from_fields(
            sizes=[3],
            features={},
            adjacency=tfgnn.Adjacency.from_indices(
                source=("users", [1, 2, 3]),
                target=("users", [0, 0, 0]),
            )),
    })
\end{python}

The data types for \pyth{tfgnn.Context}, \pyth{tfgnn.NodeSet}, \pyth{tfgnn.EdgeSet}, and \pyth{tfgnn.Adjacency} are exposed to the API for this construction to be possible. The classes verify that the ranks and shapes of the tensors are matching each other as constrained by the graph structure (i.e., in the same set they share a common prefix dimensions).

\subsubsection{Reading GraphTensors}

Graphs can be written to disk by encoding them into streams of \pyth{tf.train.Example} protos.

Graphs can be read into a TensorFlow program by decoding these \pyth{Example} protos to \pyth{GraphTensor} objects. To configure the parsing routine, the \pyth{GraphSchema} message is read and converted to a \pyth{GraphTensorSpec} object which describes the layout of the graph tensor in the TensorFlow runtime: the list of its various node sets, edge sets, and all the features associated with them and the context features. This data structure is analogous to the \pyth{tf.TensorSpec} object of TensorFlow. A \pyth{GraphTensorSpec} object is attached to all instances of \pyth{GraphTensor} and flows along with it. To read a stream of graph tensors, first create a spec, like this:

\begin{python}
import tensorflow_gnn as tfgnn

graph_schema = tfgnn.read_schema(schema_path)
graph_spec = tfgnn.create_graph_spec_from_schema_pb(graph_schema)
\end{python}

You can then use the library’s own parser to decode the \pyth{tf.train.Example} features into \pyth{GraphTensor} objects:
\begin{python}
data_paths = tf.data.Dataset.list_files(...)
ds = tf.data.TFRecordDataset(data_paths)
ds = ds.batch(batch_size)
parser_fn = functools.partial(tfgnn.parse_example, graph_spec)
ds = ds.map(parser_fn)
\end{python}

The dataset \pyth{ds} yields a stream of instances of \pyth{GraphTensor}. The parsing function ingests a batch of serialized graphs (a rank-1 tensor of strings). 

Similarly to \pyth{tf.io.parse_single_example}, there is also a function \pyth{tfgnn.parse_single_example} to parse an unbatched stream of encoded \pyth{tf.train.Example} strings with \pyth{GraphTensor} instance in them, and you can also run batch on top of that (i.e., you can batch GraphTensor instances):

\begin{python}
data_paths = tf.data.Dataset.list_files(...)
ds = tf.data.TFRecordDataset(data_paths)
parser_fn = functools.partial(tfgnn.parse_single_example, graph_spec)
ds = ds.map(parser_fn)
ds = ds.batch(batch_size)
\end{python}

Typically this step is followed by a feature engineering step that normalizes, embeds and concatenates the various features into a single embedding for each set, and then batches multiple subgraphs into modified batched \pyth{GraphTensor} instances.

The containers can then be used to pick out various tensors and build a model, or simply to inspect their contents:

\begin{python}
for graph in ds.take(10):
  tensor = graph.node_sets["item"]["category"]
  print(tensor)
\end{python}

You will go through important details of batching and flattening to a single graph below.

\subsection{Broadcast and pool operations}

Consider how the message passing API could help us to find the total user spending on purchased items.
As a preparation step, let's calculate for each "item" the latest price value and materialize the result to the graph tensor.

\begin{python}
item_features = graph.node_sets["items"].get_features_dict()
item_features["latest_price"] = item_features["price"][:, :1].values
graph = graph.replace_features(node_sets={"items": item_features})
print(graph.node_sets["items"])
\end{python}

User spending can now be calculated by first computing purchase prices by broadcasting item latest prices to the "purchase" edges.
The total user spendings are computed by sum-pooling purchase prices to their users.
\begin{python}
purchase_prices = tfgnn.broadcast_node_to_edges(
    graph, "purchased", tfgnn.SOURCE, feature_name="latest_price")
total_user_spendings = tfgnn.pool_edges_to_node(
    graph,
    "purchased",
    tfgnn.TARGET,
    reduce_type="sum",
    feature_value=purchase_prices)
print(total_user_spendings)
\end{python}
Note that API allows to reference feature values either by their name and by their values. For latter it is not required that
feature exists in the graph tensor as soon as its shape is correct.

Message passing is also supported between the graph context and any node set or edge set. As an example, let's compare individual user spendings to the maximum amount spent by any users. The code below first max-pool all individual user spendings and then broadcast them back to "users" to compute fractions.
\begin{python}
max_amount_spend = tfgnn.pool_nodes_to_context(
    graph, "users", feature_value=total_user_spendings,
    reduce_type="max")
max_amount_spend = tfgnn.broadcast_context_to_nodes(
    graph, "users", feature_value=max_amount_spend)

print(total_user_spendings / max_amount_spend)
\end{python}

\subsection{Implementing GATv2}
\label{appendix:gatv2}
The following code snippet shows the essence of how GATv2~\citep{brody2022gatv2} has been implemented in \gnn. It performs attention from many senders to one receiver.

In the simplest case, the convolution is applied to an edge set whose target nodes are the receivers, and each receiver attends to the adjacent source nodes. However, \gnn\ lets you configure this is many ways: (1)~The roles of source and target nodes can be swapped by setting the \pyth{receiver_tag}; for example, consider how an edge set of hyperlinks between HTML docs can reasonably used in either direction. (2)~The sender value can be supplied by the neighbor node, the connecting edge, or both; this is controlled by the \pyth{sender_*_feature} args, at least one of which must not be \pyth{None}. (3)~Attention can also happen with \pyth{receiver_tag=tfgnn.CONTEXT} and either all nodes or all edges in the respective graph component as senders. This comes in handy for attention in the context update of a Graph Network~\citep{battaglia18graphnet} or for readout of a graph-level feature by smart pooling of node or edge states.

All those case distinctions are handled by the superclass, which passes the suitable broadcast and pool ops as arguments into the actual \pyth{GATv2Conv.convolve()} method.

\lstset{style=mypython,basicstyle=\ttfamily\scriptsize}
\begin{lstlisting}
class GATv2Conv(tfgnn.keras.layers.AnyToAnyConvolutionBase):

  def __init__(self, *,
               num_heads, per_head_channels,
               receiver_tag=None,  # SOURCE, TARGET or CONTEXT.
               receiver_feature=tfgnn.HIDDEN_STATE,  # Required.
               sender_node_feature=tfgnn.HIDDEN_STATE,  # Set None to disable.
               sender_edge_feature=None,  # Set tfgnn.HIDDEN_STATE to enable.
               attention_activation="leaky_relu", activation="relu",
               edge_dropout=0., **kwargs):
    # Save initializer args.
    super().__init__(
        receiver_tag=receiver_tag, receiver_feature=receiver_feature,
        sender_node_feature=sender_node_feature,
        sender_edge_feature=sender_edge_feature,
        extra_receiver_ops={"softmax": tfgnn.softmax}, **kwargs)
    self._num_heads = num_heads
    self._per_head_channels = per_head_channels
    self._edge_dropout_layer = tf.keras.layers.Dropout(edge_dropout)
    self._attention_activation = tf.keras.activations.get(attention_activation)
    self._activation = tf.keras.activations.get(activation)
    # Create the transformations for the query input in all heads.
    self._w_query = tf.keras.layers.Dense(per_head_channels * num_heads)
    # Create the transformations for value input from sender nodes and edges.
    self._w_sender_node = self._w_sender_edge = None
    if self.takes_sender_node_input:
      self._w_sender_node = tf.keras.layers.Dense(per_head_channels * num_heads)
    if self.takes_sender_edge_input:
      self._w_sender_edge = tf.keras.layers.Dense(
          per_head_channels * num_heads,
          use_bias=self._w_sender_node is None)  # Avoid two biases.
    # Create the transformation to attention scores.
    self._attention_logits_fn = tf.keras.layers.experimental.EinsumDense(
        "...ik,ki->...i", output_shape=(num_heads,))

  def convolve(self, *,
               sender_node_input, sender_edge_input, receiver_input,
               broadcast_from_sender_node, broadcast_from_receiver,
               pool_to_receiver, extra_receiver_ops, training):
    # Form the attention query for each head.
    query = broadcast_from_receiver(self._split_heads(self._w_query(
        receiver_input)))
    # Form the attention value for each head.
    value_terms = []
    if sender_node_input is not None:
      value_terms.append(broadcast_from_sender_node(
          self._split_heads(self._w_sender_node(sender_node_input))))
    if sender_edge_input is not None:
      value_terms.append(
          self._split_heads(self._w_sender_edge(sender_edge_input)))
    value = tf.add_n(value_terms)
    # Compute the attention coefficients.
    attention_features = self._attention_activation(query + value)
    logits = tf.expand_dims(self._attention_logits_fn(attention_features), -1)
    attention_coefficients = extra_receiver_ops["softmax"](logits)
    attention_coefficients = self._edge_dropout_layer(attention_coefficients)
    # Apply the attention coefficients to the transformed query.
    messages = value * attention_coefficients
    pooled_messages = pool_to_receiver(messages, reduce_type="sum")
    # Apply the nonlinearity.
    return self._activation(self._merge_heads(pooled_messages))

  # The following helpers map forth and back between tensors with...
  #  - a separate heads dimension: shape [..., num_heads, channels_per_head],
  #  - all heads concatenated:    shape [..., num_heads * channels_per_head].
  def _split_heads(self, tensor):
    extra_dims = tensor.shape[1:-1]  # Possibly empty.
    new_shape = (-1, *extra_dims, self._num_heads, self._per_head_channels)
    return tf.reshape(tensor, new_shape)

  def _merge_heads(self, tensor):
    extra_dims = tensor.shape[1 : -2]  # Possibly empty.
    merged_dims = tensor.shape[-2:]
    new_shape = (-1, *extra_dims, merged_dims.num_elements())
    return tf.reshape(tensor, new_shape)
\end{lstlisting}

\subsection{Orchestration}

The following code snippet demonstrates a typical use of the \textit{Orchestrator}. Quick start users
need specify only:

\begin{enumerate}
    \item Training---and optional validation---dataset(s),
    \item Optional feature processing callbacks,
    \item A base Graph Neural Network. (A \pyth{tf.keras.Model} that both accepts and returns a \pyth{GraphTensor}),
    \item A \pyth{runner.Task}.
\end{enumerate}

The snippet assumes a toy \pyth{GraphSchema}. The provided training---and optional validation---dataset(s) must contain either (\rnum{1}) serialized
\pyth{tf.train.Example} protocol buffers to be parse by this schema or (\rnum{2}) \pyth{GraphTensor} objects compatible with this schema. (As checked by \pyth{tfgnn.check_compatible_with_schema_pb}.) The toy \pyth{GraphSchema}:

\lstset{style=mypython,basicstyle=\ttfamily\scriptsize}
\begin{lstlisting}
context {
  features {
    key: "ni"
    value: {
      dtype: DT_INT64
      shape { dim { size: 1 } }
    }
  }
}
node_sets {
  key: "coconut"
  value {
  }
}
node_sets {
  key: "grail"
  value {
    features {
      key: "holiness"
      value {
        dtype: DT_INT64
        shape { dim { size: 1 } }
      }
    }
  }
}
node_sets {
  key: "swallow"
  value {
    features {
      key: "speed"
      value {
        dtype: DT_FLOAT
        shape { dim { size: 1 } }
      }
    }
  }
}  
edge_sets {
    ...
}
edge_sets {
    ...
}
edge_sets {
    ...
}
edge_sets {
    ...
}
\end{lstlisting}

And \textit{orchestration}:

\lstset{style=mypython,basicstyle=\ttfamily\scriptsize}
\begin{lstlisting}
import tensorflow as tf
import tensorflow_gnn as tfgnn

from tensorflow_gnn import runner

gtschema = tfgnn.read_schema(".../schema.pbtxt")
gtspec = tfgnn.create_graph_spec_from_schema_pb(gtschema)

train_ds_provider = runner.TFRecordDatasetProvider(".../train*")

# Extract labels from the graph context, importantly: this lambda
# matches the `runner.GraphTensorProcessorFn` protocol.
def extract_label_fn(gt):
    return gt, gt.context["ni"]

# Process node features, importantly: this callback is executed as
# a part of a `tf.data.Dataset`. (Execution takes place on CPU and
# not any accelerator.)
def node_sets_processor_fn(node_set, *, node_set_name):
  if node_set_name == "coconut":
    return {"empty": tfgnn.keras.layers.MakeEmptyFeature(node_set)}
  if node_set_name == "grail":
    return {"holiness": tf.keras.layers.Hashing(153)(node_set["holiness"])}
  if node_set_name == "swallow":
    return {"speed": tf.math.log1p(node_set["speed"])}
  
# Set initial node states, importantly: this callback is executed
# as a part of modeling. (Learnable weights are included in back
# propagation.)
def initial_node_states_fn(node_set, *, node_set_name)
  if node_set_name == "coconut":
    return node_set["empty"]
  if node_set_name == "grail":
    return tf.keras.layers.Embedding(153, 16)(node_set["holiness"])
  if node_set_name == "swallow":
    return tf.keras.layers.Dense(16)(node_set["speed"])

def model_fn(gtspec):
  # simple_gnn is a function: Callable[[tfgnn.GraphTensorSpec], tf.keras.Model]. Where
  # the returned model both takes and returns a scalar `GraphTensor`
  # for its inputs and outputs.
  return tf.keras.Sequential([
      tfgnn.keras.layers.MapFeatures(node_sets_fn=initial_node_states_fn),
      simple_gnn(gtspec),
  ])
    
# Binary classification by the root node.
task = runner.RootNodeBinaryClassification(node_set_name="grail")

trainer = runner.KerasTrainer(
    strategy=tf.distribute.TPUStrategy(),
    model_dir=".../monty/",
    # len(train_ds_provider.get_dataset(...)) == 8191 and global_batch_size == 128.
    steps_per_epoch=8191 // 128,
    restore_best_weights=False)  

runner.run(
    train_ds_provider=train_ds_provider,
    train_padding=runner.FitOrSkipPadding(gtspec, train_ds_provider),
    model_fn=model_fn,
    optimizer_fn=tf.keras.optimizers.Adam,
    epochs=4,
    trainer=trainer,
    task=task,
    gtspec=gtspec,
    global_batch_size=128,
    feature_processors=[
        extract_label_fn,  # Extract any labels first.
        tfgnn.keras.layers.MapFeatures(node_sets_fn=node_sets_processor_fn),
    ])
\end{lstlisting}

\subsection{Case Study: OGBN-MAG}

Full definitions for files used in the case study on OGBN-MAGG\footnote{\url{https://ogb.stanford.edu/docs/nodeprop/\#ogbn-mag}} (Section \ref{sec:case_study}) are provided here for completeness.

\subsubsection{Graph Schema}

This is the protocol buffer definition of the Graph Schema used for OGBN-MAG from Figure \ref{fig:case_study_schema}.

\begin{lstlisting}
node_sets {
  key: "author"
  value {
    features {
      key: "#id"
      value {
        dtype: DT_STRING
      }
    }
    metadata {
      filename: "nodes-author.tfrecords@15"
      cardinality: 1134649
    }
  }
}
node_sets {
  key: "field_of_study"
  value {
    features {
      key: "#id"
      value {
        dtype: DT_STRING
      }
    }
    metadata {
      filename: "nodes-field_of_study.tfrecords@2"
      cardinality: 59965
    }
  }
}
node_sets {
  key: "institution"
  value {
    features {
      key: "#id"
      value {
        dtype: DT_STRING
      }
    }
    metadata {
      filename: "nodes-institution.tfrecords"
      cardinality: 8740
    }
  }
}
node_sets {
  key: "paper"
  value {
    features {
     key: "#id"
      value {
        dtype: DT_STRING
      }
    }
    features {
      key: "feat"
      value {
        dtype: DT_FLOAT
        shape {
          dim {
            size: 128
          }
        }
      }
    }
    features {
      key: "labels"
      value {
        dtype: DT_INT64
        shape {
          dim {
            size: 1
          }
        }
      }
    }
    features {
      key: "year"
      value {
        dtype: DT_INT64
        shape {
          dim {
            size: 1
          }
        }
      }
    }
    metadata {
      filename: "nodes-paper.tfrecords@397"
      cardinality: 736389
    }
  }
}
edge_sets {
  key: "affiliated_with"
  value {
    source: "author"
    target: "institution"
    metadata {
      filename: "edges-affiliated_with.tfrecords@30"
      cardinality: 1043998
    }
  }
}
edge_sets {
  key: "cites"
  value {
    source: "paper"
    target: "paper"
    metadata {
      filename: "edges-cites.tfrecords@120"
      cardinality: 5416271
    }
  }
}
edge_sets {
  key: "has_topic"
  value {
    source: "paper"
    target: "field_of_study"
    metadata {
      filename: "edges-has_topic.tfrecords@226"
      cardinality: 7505078
    }
  }
}
edge_sets {
  key: "writes"
  value {
    source: "author"
    target: "paper"
    metadata {
      filename: "edges-writes.tfrecords@172"
      cardinality: 7145660
    }
  }
}

\end{lstlisting}

\subsubsection{Generated Sample Spec} 

This is the protocol buffer generated for input to a sampling pipeline from the code snippet in Figure \ref{alg:samplingspecbuilder}.

\begin{lstlisting}
seed_op {
  op_name: "SEED->paper"
  node_set_name: "paper"
}
sampling_ops {
  op_name: "paper->paper"
  input_op_names: "SEED->paper"
  edge_set_name: "cites"
  sample_size: 32
  strategy: RANDOM_UNIFORM
}
sampling_ops {
  op_name: "(paper->paper|SEED->paper)->author"
  input_op_names: "paper->paper"
  input_op_names: "SEED->paper"
  edge_set_name: "written"
  sample_size: 8
  strategy: RANDOM_UNIFORM
}
sampling_ops {
  op_name: "author->institution"
  input_op_names: "(paper->paper|SEED->paper)->author"
  edge_set_name: "affiliated_with"
  sample_size: 16
  strategy: RANDOM_UNIFORM
}
sampling_ops {
  op_name: "author->paper"
  input_op_names: "(paper->paper|SEED->paper)->author"
  edge_set_name: "writes"
  sample_size: 16
  strategy: RANDOM_UNIFORM
}
sampling_ops {
  op_name: "(author->paper|SEED->paper|paper->paper)->field_of_study"
  input_op_names: "author->paper"
  input_op_names: "SEED->paper"
  input_op_names: "paper->paper"
  edge_set_name: "has_topic"
  sample_size: 16
  strategy: RANDOM_UNIFORM
}
\end{lstlisting}

\subsubsection{Hyper-parameter optimization}
\label{appendix:hyper-params-optimization}

In their Vizier study, our Googler sets up the following hyperparameter search problem: 
\begin{enumerate}
    \item `message\_dim`: the dimension of hidden states for each node. Discrete Search space: `[32, 64, 128, 256, 512]`  
    \item `reduce\_type`: the strategy to pool the messages from edges to receiver nodes.  Categorical Search space: `[“sum”, “mean”], 
    \item `l2\_regularization`: the coefficient of l2 regularization for weights and biases. Continuous (Log) Search Space: `[1e-6, 1e-4].
    \item `dropout`: the dropout probability. Discrete Search space: `[0.1, 0.2, 0.3]`
    \item `use\_layer\_normalization`: Flag to determine whether to utilize layer normalization when updating the node states. Boolean Search space: `[True, False]`  
\end{enumerate}

They use the Adam optimizer with a cosine decay learning rate schedule.

\textbf{Results.} The top-3 performing models on the validation set had `message\_dim = 256`, `reduce\_type = “sum”`, `dropout = 0.2`, `use\_layer\_normalization = True` while the `l2\_regularization` varied in the range [1e-6, 4e-6].

\subsubsection{A minimal usage of the TF-GNN Orchestration Layer}
\label{appendix:minimal-runner-ogbnmag}

This illustrates how to train the model defined in Section \ref{alg:case_study_mympnn} using the Orchestrator.

\begin{lstlisting}
from tensorflow_gnn import runner

train_ds_provider = runner.TFRecordDatasetProvider("/home/...")
feature_processors = [extract_labels, feature_mapping]  # See appendix.

task = runner.RootNodeMulticlassClassification(
    node_set_name="paper", num_classes=349)

trainer = runner.KerasTrainer(
   strategy=tf.distribute.MirroredStrategy(), model_dir="/tmp/my_model",
   steps_per_epoch=num_training_examples // global_batch_size,
   restore_best_weights=False)

optimizer_fn = functools.partial(
    tf.keras.optimizers.Adam,
    learning_rate=tf.keras.optimizers.schedules.CosineDecay(...))

runner.run(
   train_ds_provider=train_ds_provider,
   gtspec=tfgnn.create_graph_spec_from_schema_pb(graph_schema),
   feature_processors=feature_processors,
   task=task, model_fn=model_fn, # Model as defined above.
   trainer=trainer, optimizer_fn=optimizer_fn,
   epochs=20, global_batch_size=global_batch_size)
\end{lstlisting}


\end{document}

%% file: images/graph_schema_example.tex
\begin{figure*}[t!]
    \centering
    \vspace{-5pt}
     \centering
     \begin{subfigure}[b]{0.48\linewidth}
         \centering
         \includegraphics[width=\linewidth]{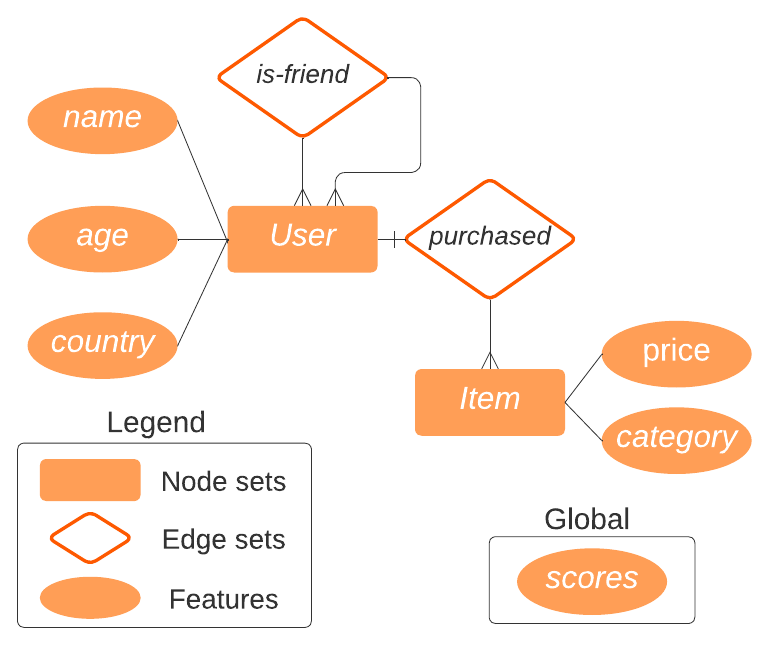}
         \caption{A graph schema.}
         \label{fig:example_schema_fig}
     \end{subfigure}
     \hfill
     \begin{subfigure}[b]{0.5\linewidth}
         \centering
        \resizebox{\linewidth}{!}{%
        \begin{tikzpicture}
        \node[draw=black, ellipse, fill=gooPurple, text width=2.5cm, align=center] (food) at (0,0) {\small food \\ {\tiny [\$22.34, \$23.42, \$12.99]}};
        \node[below right=0cm of food] {0};
        \node[draw=black, ellipse, fill=gooPurple, text width=2cm, align=center] (show ticket) at (-1,-1.5) {\small show ticket \\ {\tiny [\$27.99, \$34.50]}};
        \node[below right=0cm of show ticket] {1};
        \node[draw=black, ellipse, fill=gooPurple, text width=1.5cm, align=center] (shoes) at (-0.3,-3) {\small shoes \\ {\tiny [\$89.99]}};
        \node[below right=0cm of shoes] {2};
        \node[draw=black, ellipse, fill=gooPurple, text width=2cm, align=center] (book) at (-0.45,-4.5) {\small book \\ {\tiny [\$24.99, \$45.00]}};
        \node[below right=0cm of book] {3};
        \node[draw=black, ellipse, fill=gooPurple, text width=1.5cm, align=center] (flight) at (-1,-6) {\small flight \\ {\tiny [\$350.00]}};
        \node[below right=0cm of flight] {4};
        \node[draw=black, ellipse, fill=gooPurple, text width=2.5cm, align=center] (groceries) at (-0.25,-7.5) {\small groceries \\ {\tiny [\$45.13, \$79.80, \$12.35]}};
        \node[below right=0.035cm of groceries] {5};
        
        \node[draw=black, ellipse, fill=gooBlue, text width=1.5cm, align=center] (jeorg) at (7,-0.5) {\small Jeorg, 32, "uk"};
        \node[below right=0.075cm of jeorg] {1};
        \node[draw=black, ultra thick, ellipse, fill=gooBlue, text width=1.5cm, align=center] (shawn) at (5,-2.5) {\small Shawn, 24, "usa"};
        \node at ([shift=({340:1.3cm})]shawn) {0};
        \node[draw=black, ellipse, fill=gooBlue, text width=1.5cm, align=center] (yumiko) at (8, -4.5) {\small Yumiko, 27, "japan"};
        \node[below right=0.075cm of yumiko] {2};
        \node[draw=black, ellipse, fill=gooBlue, text width=1.5cm, align=center] (sophie) at (6, -6) {\small Sophie, 38, "france"};
        \node[below right=0.075cm of sophie] {3};
        
        \draw [->, thick] (food) edge (jeorg) (show ticket) edge (jeorg) (shoes) edge (shawn) (book) edge (shawn) (flight) edge (yumiko) (groceries) edge (shawn) (groceries) edge (sophie);
        
        \draw [->, thick, dashed] (sophie) edge (shawn) (yumiko) edge (shawn) (jeorg) edge (shawn);
        
        \node at ([shift=({120:1.3cm})]food) {\emph{items}};
        \node at ([shift=({90:1cm})]jeorg) {\emph{users}};
        \node at ([shift=({165:4cm})]jeorg) {\emph{purchased}};
        \node[below=0.35cm of jeorg] {\emph{is-friend}};
        
        \node[draw=black, rectangle, rounded corners=5pt, fill=gooGreen, text width=2.5cm, align=center] (scores) at (8, -8) {\small scores \\ {\tiny [0.45, 0.98, 0.10, 0.25]}};
        \end{tikzpicture}
    }
         \caption{A heterogeneous graph conforming to the schema.}
         \label{fig:example_graph}
     \end{subfigure}
     \hfill

    \caption{The recommending system example in the text uses the GraphSchema visualized in part~(a). Part~(b) shows a possible graph with the node sets, connecting edge sets, and features prescribed by the schema.
    }
    \label{fig:graph_schema_example_graph}
\end{figure*}